\pdfoutput=1
\documentclass{article} 
\usepackage{iclr2023_conference,times}

\usepackage{amsmath,amsfonts,bm}

\def\eqref#1{equation~\ref{#1}}

\def\1{\bm{1}}

\DeclareMathAlphabet{\mathsfit}{\encodingdefault}{\sfdefault}{m}{sl}
\SetMathAlphabet{\mathsfit}{bold}{\encodingdefault}{\sfdefault}{bx}{n}

\usepackage[hidelinks]{hyperref}
\usepackage{url}
\usepackage[pdftex]{graphicx}
\usepackage{longtable}
\usepackage{booktabs}
\usepackage{multirow}
\usepackage{tabularx}
\usepackage{comment}
\usepackage{threeparttable}
\usepackage{mathtools}
\usepackage{ulem}

\title{Prompt-Based Metric Learning \\ for Few-Shot NER}

\iclrfinalcopy

\author{
Yanru Chen$^1$, Yanan Zheng$^1$, Zhilin Yang$^{123}$\thanks{Corresponding author.} \\
$^1$Tsinghua University, $^2$Shanghai Artificial Intelligence Laboratory, $^3$Shanghai Qizhi Institute \\
\texttt{\{achen.cyanr.qaq, zyanan93\}@gmail.com} \\
\texttt{zhiliny@tsinghua.edu.cn}
}

\newcommand\name{ProML}
\newcommand\ours{\name\space}
\newcommand{\std}{\scriptsize$\pm$}
\newcommand\mixrate{$\rho$\space}

\begin{document}

\maketitle

\begin{abstract}
Few-shot named entity recognition (NER) targets generalizing to unseen labels and/or domains with few labeled examples.
Existing metric learning methods compute token-level similarities between query and support sets, but are not able to fully incorporate label semantics into modeling.
To address this issue, we propose a simple method to largely improve metric learning for NER: 1) multiple prompt schemas are designed to enhance label semantics; 
2) we propose a novel architecture to effectively combine multiple prompt-based representations.
Empirically, our method achieves new state-of-the-art (SOTA) results under 16 of the 18 considered settings, substantially outperforming the previous SOTA by an average of 8.84\% and a maximum of 34.51\% in relative gains of micro F1. Our code is available at \url{https://github.com/AChen-qaq/ProML}.

\end{abstract}

\section{Introduction}

Named entity recognition (NER) is a key natural language understanding task that extracts and classifies named entities mentioned in unstructured texts into predefined categories. Few-shot NER targets generalizing to unseen categories by learning from few labeled examples.

Recent advances for few-shot NER use metric learning methods which compute the token-level similarities between the query and the given support cases.
\citet{protobert} proposed to use prototypical networks that learn prototypical representations for target classes. Later, this method was introduced to few-shot NER tasks \citep{DBLP:conf/sac/FritzlerLK19,hou-etal-2020-shot}. 
~\citet{yang-katiyar-2020-simple} proposed StructShot, which uses a pretrained language model as a feature extractor and performs viterbi decoding at inference.
~\citet{das-etal-2022-container} proposed CONTaiNER based on contrastive learning. This approach optimizes an objective that characterizes the distance of Gaussian distributed embeddings under the metric learning framework.

Despite the recent efforts, there remain a few critical challenges for few-shot NER. First of all, as mentioned above, metric learning computes token-level similarities between the query and support sets. However, the architectures used for computing similarities in previous work are agnostic to the labels in the support set. This prevents the model from fully leveraging the label semantics of the support set to make correct predictions. Second, while prompts have been demonstrated to be able to reduce overfitting in few-shot learning \citep{PET-paper}, due to a more complex sequence labeling nature of NER, the optimal design of prompts remains unclear for few-shot NER.

\begin{figure*}
    \centering
    \includegraphics[width=\textwidth]{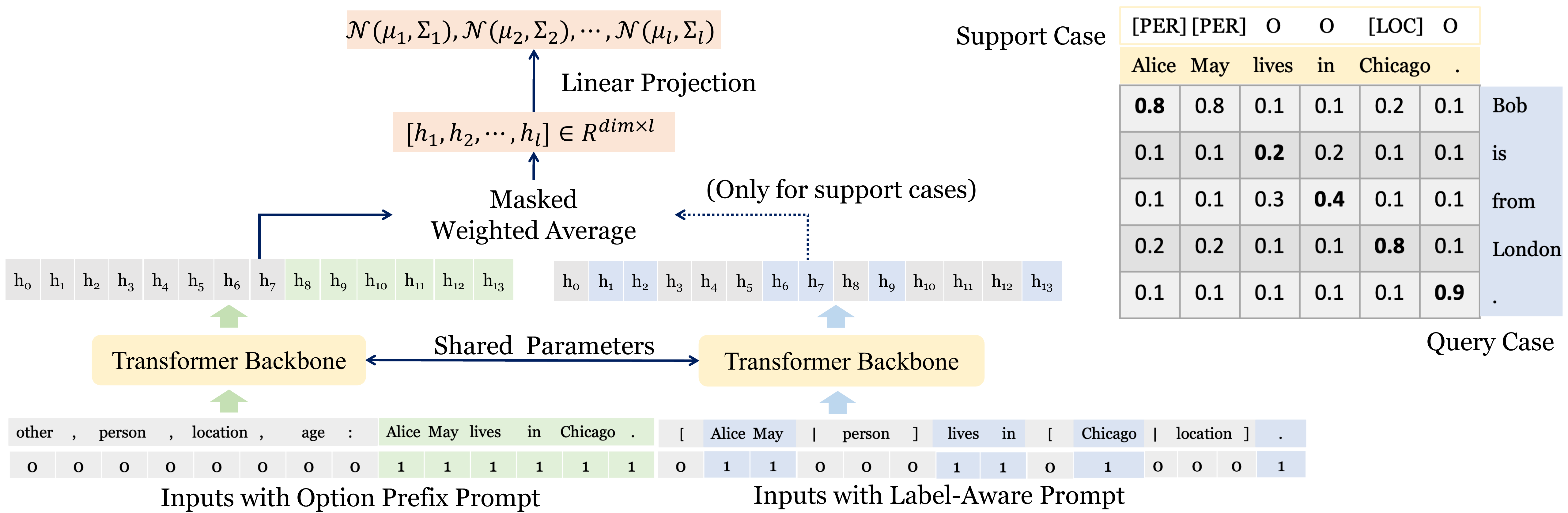}
    \caption{\small An overview of the architecture of our proposed \ours. The prompts associated with the input sequence are passed through a transformer backbone to obtain intermediate representations. A masked weighted average is then applied to produce token-level representations. Following ~\citet{das-etal-2022-container}, Gaussian embeddings for each token are produced using linear projections. The similarity scores between query tokens and support tokens are then computed according to the distance metric.}
    \label{fig:overview}
\end{figure*}

In light of the above challenges, we explore a better architecture that allows using prompts to fully leverage the label semantics. We propose a simple method of Prompt-based Metric Learning (\name) for few-shot NER, as shown in Figure~\ref{fig:overview}. Specifically, we introduce a special class of prompts, which is called the mask-reducible prompts. By performing a masked weighted average over the representations obtained from multiple prompts, our method accepts multiple choices of prompts as long as they are mask-reducible. These prompts improve label efficiency by inserting semantic annotations into the text inputs. As instantiations of this framework, we design an option prefix prompt to provide the model with the candidate label options, and a label-prefix prompt to associate each entity with its entity type in the input. 


In our experiments, we find that using multiple prompts with the masked weighted average is effective for few-shot NER.  Empirically, our method achieves new state-of-the-art (SOTA) results under 16 of the 18 considered settings, substantially outperforming the previous SOTA by an average of 8.84\% and a maximum of 34.51\% in relative gains of micro F1.

\section{Related Work}
\paragraph{Few-Shot NER.} 
Few-shot NER targets generalizing to unseen categories by learning from few labeled examples.
Noisy supervised methods~\citep{huang2020few} perform supervised pre-training over large-scale noisy web data such as WiNER~\citep{ghaddar2017winer}. Self training methods~\citep{wang2021meta} perform semi-supervised training over a large amount of unlabelled data. Alternative to these data-enhancement approaches, metric learning based methods have been widely used for few-shot NER~\citep{DBLP:conf/sac/FritzlerLK19, yang-katiyar-2020-simple, das-etal-2022-container}. Recently, prompt-based methods~\citep{ma2021template, cui-etal-2021-template, lee-etal-2022-good} are proposed for few-shot NER as well. To introduce more fine-grained entity types in few-shot NER, a large-scale human-annotated dataset Few-NERD~\citep{ding-etal-2021-nerd} was proposed. ~\citet{Ma2022DecomposedMF, Wang2022AnES} formulate NER task as a span matching problem and decompose it to several procedures. \citet{Ma2022DecomposedMF} decomposed the NER task into span detection and entity typing, and they separately train two models and finetune them on the test support set, achieving SOTA results on Few-NERD~\citep{ding-etal-2021-nerd}.
Different from the above related works, our approach is a general framework of using prompts for token-level metric learning problems.


\paragraph{Meta Learning.}
The idea of meta learning was first introduced in few-shot classification tasks for computer vision, attempting to learn from a few examples of unseen classes. Since then metric-based methods have been proposed, such as matching networks~\citep{vinyals2016matching} and Prototypical networks~\citep{protobert}, which basically compute similarities according to the given support set, learn prototypical representations for target classes, respectively.
It has been shown that these methods also enable few-shot learning for NLP tasks such as text classification~\citep{bao2019few, geng2019induction}, relation classification~\citep{han2018fewrel}, named entity recognition~\citep{DBLP:conf/sac/FritzlerLK19, yang-katiyar-2020-simple, das-etal-2022-container}, and machine translation~\citep{gu-etal-2018-meta}. Our approach also falls into the category of metric-based meta learning and outperforms previous work on NER with an improved architecture.

\paragraph{Label Semantics for NER.}
There have been some approaches that make use of label semantics~\citep{ma-etal-2022-label, hou-etal-2020-shot}.~\citet{hou-etal-2020-shot} propose a CRF framework with label-enhanced representations based on the architecture of~\citet{yoon2019tapnet}.
However, they mainly focus on slot tagging tasks while their performance on NER tasks is poor. ~\citet{ma-etal-2022-label} introduce label semantics by aligning token representations with label representations. Both of them only use label semantics for learning better label representations.
In contrast, our approach incorporates label semantics into the inputs so that the model is able to jointly model the label information and the original text samples. This makes the similarity scores dependent on the support set labels and is particularly crucial for metric learning. Our experiments also verify the advantages of our approach compared to previous work using labels semantics.

\paragraph{Prompt-Based Approaches for NER.}
With the emergence of prompt-based methods in NLP research, very recently, some prompt-based approaches for few-shot NER have been proposed~\citep{cui-etal-2021-template, lee-etal-2022-good, ma2021template}.
However, they use prompts to help with the label predictions based on classification heads instead of metric learning. Moreover, some of these methods require searching for templates~\citep{cui-etal-2021-template}, good examples~\citep{lee-etal-2022-good}, or label-aware pivot words~\citep{ma2021template}, which makes the results highly dependent on the search quality. Different from these methods, our approach does not rely on a search process. More importantly, another key difference is that we employ prompting in the setting of metric learning.


\section{Task Definition}
\subsection{Few-shot NER}
Named entity recognition (NER) is a sequence labeling task\footnote{There also exist other formulations such as span prediction or question answering.}. Formally, for a sentence $\mathbf{x}$ consisting of $n$ tokens $\mathbf{x}=[x_1, x_2, \cdots, x_n]$, there is a corresponding ground-truth label sequence $\mathbf{y}=[y_1, y_2, \cdots, y_n]$ where each $y_i$ is an encoding of some label indicating the entity type for token $x_i$. Then a collection of these $(\mathbf{x, y})$ pairs form a dataset $\mathcal{D}$. After training on the training dataset $\mathcal{D_S}$, the model is required to predict labels for sentences from the test dataset $\mathcal{D_T}$.

Different from the standard NER task, the few-shot NER setting consists of a meta training phase and a test phase. At the meta training phase, the model trains on a training dataset $\mathcal{D_S}$. At the test phase, for various test datasets $\{\mathcal{D_T}^{(j)}\}$, with only few labeled samples, the model is required to perform quick adaptions. In this paper, we mainly focus on two evaluation protocols and two task formulations which will be explained as follows.
\subsection{Evaluation protocols}
Following \citet{ding-etal-2021-nerd,ma-etal-2022-label}, we summarize two evaluation protocols as follows.
\paragraph{Episode Evaluation} An episode, or a task, is defined as a pair of one support set and one query set $(\mathcal{S, Q})$ each consisting of sentences downsampled from the test set. For an $N$-way $K$-shot downsampling scheme, there are $N$ labels among the support set $\mathcal{S}$ where each label is associated with $K$ examples. The query set $\mathcal{Q}$ shares the same label set with the support set. Based on the support set, the model is required to predict labels for the query set. To perform an episode evaluation, a collection of $T$ episodes $\{(\mathcal{S}_t, \mathcal{Q}_t)\}_{t=1}^T$ are prepared. The evaluation results are computed within each episode and are averaged over all $T$ episodes.
\paragraph{Low-resource Evaluation} Different from the few-shot episode evaluation, low-resource evaluation aims to directly evaluate the model on the whole test set. For a test dataset $\mathcal{D_T}$ with a label set $\mathcal{C_T}$, a support set $\mathcal{S}$ associated with the labels from $\mathcal{C_T}$ is constructed by $K$-shot downsampling such that each label has $K$ examples in $\mathcal{S}$. Based on the support set $\mathcal{S}$, the model is required to predict labels for the query set which is the rest of the test set $\mathcal{D_T}$. To perform a low-resource evaluation, $T$ different runs of support set sampling are run and averaged.

\subsection{Task formulation}
Following~\citet{yang-katiyar-2020-simple}, we formulate few-shot NER tasks in the following two ways.
\paragraph{Tag-Set Extension}
To mimic the scenario that new classes of entities emerge in some domain, \citet{yang-katiyar-2020-simple} propose the tag-set extension formulation. Starting with a standard NER dataset $(\mathcal{D}_{train}, \mathcal{D}_{test})$ with label set $\mathcal{C}$, they split $\mathcal{C}$ into $d$ parts, namely $\mathcal{C}_1, \mathcal{C}_2, \cdots, \mathcal{C}_d$. Then for each label split $\mathcal{C}_i$, a train set $\mathcal{D}_{train}^{(i)}$ is constructed from $\mathcal{D}_{train}$ by masking the labels in $\mathcal{C}_i$ to $O$ (representing non-entities), and the corresponding test set $\mathcal{D}_{test}^{(i)}$ is constructed from $\mathcal{D}_{test}$ by masking the labels in $\mathcal{C}\setminus \mathcal{C}_i$ to $O$. 

\paragraph{Domain Transfer}
Another task formulation is the domain transfer setting. Let $\mathcal{D_S}$ be a training set of a standard NER task, and let $\{\mathcal{D_T}^{(i)}\}$ be the test sets of standard NER tasks but from a different domain. The training set $\mathcal{D_S}$ is referred to as a source domain, and the test sets $\{\mathcal{D_T}^{(i)}\}$ constitute various target domains. In this setting, there may exist some overlapping entity classes between the source and target domains, but 
due to the domain gaps, it is still considered a few-shot setting.

Note that the task formulation is independent of the evaluation protocol, and different combinations will be considered in our experiments.


\section{Method}

\subsection{Prompt Schemas}

Motivated by existing prompt-based methods~\citep{DBLP:journals/corr/abs-2107-13586,paolini2021structured} and the metric learning framework, our \ours provides label semantics by introducing prompts to metric learning models.
We proposed a simple yet effective prompt class called the ``mask-reducible prompts''. Through this class of prompts, we can provide flexible prompts to the model which is consistent with metric learning methods that use token-level similarities as the metric. Starting with this schema, we will introduce two prompts that are used in \ours, the option-prefix prompt and the label-aware prompt.
\subsubsection{Mask-Reducible Prompts}


Suppose the raw input sequence is $\mathbf{x}=[x_1, x_2, \cdots, x_l]$. Let $f_{prompt}$ be a prompt function that maps $\mathbf{x}$ to the prompted result $\mathbf{x'}$. We call this $f_{prompt}$ is a mask-reducible prompt function if for all $\mathbf{x}$ and its prompted result $\mathbf{x'}=f_{prompt}(\mathbf{x})$, there exists a mask $\mathbf{m}\in [0, 1]^{|\mathbf{x'}|}$ such that $\mathbf{x'}[\mathbf{m}==1]=\mathbf{x}$. Intuitively, this means there is only some insertions in the prompt construction so that we can revert $\mathbf{x'}$ back to $\mathbf{x}$ through a simple masking operation. The corresponding prompt of $f_{prompt}$ is called a mask-reducible prompt.

Let the collection of mask-reducible prompt functions be a function class $\mathcal{F}$. We can make use of the nice property of this function class to extract token-level features for the whole input sequence. 

Given a length preserving sequence-to-sequence encoder $Enc(\mathbf{x};\theta)$, a sequence of input tokens $\mathbf{x}$, and a mask-reducible prompt function $f_{prompt}\in \mathcal{F}$, we first construct the prompted result $\mathbf{x'}=f_{prompt}(\mathbf{x})$, then pass the sequence $\mathbf{x'}$ through the encoder to get representations $\mathbf{h'} = Enc(\mathbf{x'};\theta)$.

Since $Enc(\cdot;\theta)$ is a length preserving encoder, the length of $\mathbf{h'}$ is exactly the same as $\mathbf{x'}$, and we can compute $\mathbf{h}=\mathbf{h'}[\mathbf{m}==1]$ to get the representation for input tokens, where $\mathbf{m}$ is the desired mask that could reduce $\mathbf{x'}$ to $\mathbf{x}$ (i.e. $\mathbf{x'}[\mathbf{m}==1]=\mathbf{x}$).

Through this process, the encoder receives the full prompts as its input while only the representations of raw input tokens are extracted.
\subsubsection{Prompt A: Option Prefix Prompts}

An option prefix prompt takes the concatenation of all annotations as an option prefix to incorporate label semantics into modeling.
Formally, for a given set of label options $\mathbf{S}=\{\mathbf{s_1}, \mathbf{s_2}, \cdots, \mathbf{s_{|S|}}\}$, we construct construct a mask-reducible prompting function $f_{A}(\mathbf{x}, \mathbf{S})$ associated with $\mathbf{S}$ using the template ``$\mathbf{s_1}, \mathbf{s_2}, \cdots, \mathbf{s_{|S|}}:\mathbf{x}$''. For example, suppose a set of label options is \texttt{\{other, person, location, age\}}, the original input sentence is \texttt{[Alice May lives in Chicago.]}, then the prompted result will be \texttt{[other, person, location, age: Alice May lives in Chicago.]}. The option prefix prompts inform the main model of which labels to predict, which can be used to learn label-dependent representations for computing the similarities. The main idea is also illustrated in Figure \ref{fig:overview}.


\subsubsection{Prompt B: Label-Aware Prompts}


A label-aware prompt appends the entity type to each entity occurrence in the input so that the model is aware of such information. While the aforementioned option prefix prompts incorporate global label information, the label-aware prompts introduce local information about each entity.
Specifically, let $f_{B}(\mathbf{x}, \mathbf{y})$ be the prompt function. Given a sequence of input tokens $\mathbf{x}$ and its ground-truth label sequence $\mathbf{y}$, for each entity $\mathbf{e}$ that occurs in $\mathbf{x}$, we obtain its corresponding label $\mathbf{E}$ from the sequence $\mathbf{y}$, and replace $\mathbf{e}$ with an label-appended version ``[$\mathbf{e} | \mathbf{E}$]'' to construct the prompted result $\mathbf{x'}=f_{B}(\mathbf{x}, \mathbf{y})$. Both the entity $\mathbf{e}$ and its label $\mathbf{E}$ are sequences of tokens. An example is given in Figure \ref{fig:overview}. The label-aware prompt is also a mask-reducible prompt because it only contains insertions. Moreover, because the label-aware prompt can be applied when the ground-truth label is available, in our few-shot learning setting, we do not apply this prompt to the query set. More details will be explained in the following descriptions of our model architecture.

Note that it is possible to design other mask-reducible prompts for NER, which will be naturally handled by our framework. In our study, we find these two prompts work well practically and use them as instantiations to demonstrate the effectiveness of our framework.




\subsection{Model and Training}
The overall architecture of \ours is shown in Figure \ref{fig:overview}. Our architecture uses a transformer backbone to encode different prompted inputs separately and employs a masked weighted average to obtain token representations, which will be elaborated as follows.



At the meta training phase, we sample mini-batches from the training set $\mathcal{D}_{train}$, where each mini-batch contains a few-shot episode $(\mathcal{S}_{train}, \mathcal{Q}_{train})$. We obtain the label set associated with the support set $\mathcal{S}_{train}$ and use a look-up dictionary to translate each label id to its natural language annotation. This leads to a set of label annotations $\mathbf{S}$.
Then for an input sequence $\mathbf{x} = [x_1, x_2, \cdots, x_l]$ and its label sequence $\mathbf{y} = [y_1, y_2, \cdots, y_l]$ from the support set $\mathcal{S}_{train}$, we collect the prompted results $\mathbf{p_A}=f_A(\mathbf{x}, \mathbf{S})$, $\mathbf{p_B}=f_B(\mathbf{x}, \mathbf{y})$ and the corresponding masks $\mathbf{m_A}$, $\mathbf{m_B}$. 
These prompted results are then passed through a pretrained language model $\operatorname{PLM}$. The average of outputs from the last four hidden layers are computed as the intermediate representations
\begin{align*}
    \mathbf{h_A} = \operatorname{PLM}(\mathbf{p_A}), \mathbf{h_B} = \operatorname{PLM}(\mathbf{p_B}).
\end{align*}





We perform a masked weighted average to obtain token representations
\begin{align*}
    \mathbf{h} = \rho \mathbf{h_A}[\mathbf{m_A}==1] + (1 - \rho) \mathbf{h_B}[\mathbf{m_B}==1],
\end{align*}

where $\rho\in (0, 1)$ is a hyperparameter.

The token representations for the query set are computed similarly. However, during both training and testing, we only use the option-prefix prompt for the query set since the ground-truth label sequence will not be available at test time. As a result, we do not perform a weighted average for the query set.
After obtaining the token representations, two projection layers $f_{\mu}, f_{\Sigma}$ are employed to produce two Gaussian embeddings, i.e., the mean and precision parameters of a $d$-dimensional Gaussian distribution $\mathcal{N}_{(\mu, \Sigma)}$ for each token in the query and support sets~\citep{das-etal-2022-container}.


Given the Gaussian embeddings for samples in both the support and query sets, we 
compute the distance metrics. Similar to CONTaiNER~\citep{das-etal-2022-container},  for a token $x_i$ from the support set $\mathcal{S}_{train}$ and a token $x'_j$ from the query set $\mathcal{Q}_{train}$, the distance between two tokens $x_i, x'_j$ is defined as the Jenson-Shannon divergence~\citep{DBLP:conf/isit/FugledeT04} of their Gaussian embeddings, i.e., 
\begin{align*}
    dist(x_i, x'_j) &= D_{JS}(\mathcal{N}_i, \mathcal{N}'_j) = \frac12 \left(D_{KL}(\mathcal N_{(\mu_i,\Sigma_i)}||\mathcal N_{(\mu'_j,\Sigma'_j)})+ D_{KL}(\mathcal N_{(\mu'_j,\Sigma'_j)}||\mathcal N_{(\mu_i,\Sigma_i)})\right),
\end{align*}
where $D_{KL}$ refers to the Kullback–Leibler divergence.

The similarity between $x_i$ and $x'_j$ is then defined as $s(x_i, x'_j)=\exp(-dist(x_i, x'_j))$.
Let $\overline{\mathcal S}_{train}, \overline{\mathcal Q}_{train}$ be collections of all tokens from sentences in $\mathcal S_{train}, \mathcal Q_{train}$.
For each $q \in \overline{\mathcal Q}_{train}$, the associated loss function is computed as
\begin{align*}
    \ell(q) = -\log \frac{\sum_{p\in\mathcal X_q} s(q,p)/|\mathcal X_q|}{\sum_{p\in \overline{\mathcal S}_{train}} s(q, p)},
\end{align*}
where $\mathcal X_q$ is defined by
$
    \mathcal X_q = \{p\in \overline{\mathcal S}_{train} | p,q \text{ have the same labels}\}.
$
The overall loss function within a mini-batch is the summation of token-level losses,
$
    L = \frac1{|\overline{\mathcal Q}_{train}|} \sum_{q\in \overline{\mathcal Q}_{train}} \ell(q).
$


\subsection{Nearest Neighbor Inference}

At test time, we compute the intermediate representations for tokens from the support and query sets just as we did during the meta training phase.
Following CONTaiNER~\citep{das-etal-2022-container}, we no longer use the projection layers $f_{\mu}, f_{\Sigma}$ at test time but directly perform nearest neighbor inference using the token representations $\mathbf{h}$. For each query token, according to the Euclidean distance in the representation space, we select its nearest neighbor among all tokens from the support set and assign the label associated with this nearest neighbor to the query token.

\begin{table*}[t]
\setlength{\tabcolsep}{1.0mm}
\centering
\caption{\small Evaluation results of \ours and 8 baseline methods in low-resource evaluation protocol for both tag-set extension and domain transfer tasks. Results with $\star$ are reported by the original paper, and those with $\dagger$ are reproduced in our experiments. We report the averaged micro-F1 score together with standard deviation. ``Onto-A'' denotes group-A set of OntoNotes dataset.
}
\resizebox{\textwidth}{!}{%
\begin{threeparttable}
    \begin{tabular}{l|ccc|cccc|c}
    \toprule[1pt]
    \multirow{2}{*}{Method} & \multicolumn{3}{c|}{Tag-Set Extension} & \multicolumn{4}{c|}{Domain Transfer} & \multirow{2}{*}{Avg.} \\
    & Onto-A & Onto-B & Onto-C & CoNLL & WNUT & I2B2 & GUM \\ 
    \midrule[1pt]
    \multicolumn{9}{c}{1-shot} \\
    \midrule[1pt]
    \multirow{1}*{ProtoBERT($\star$)}
        & 19.3\std3.9
        & 22.7\std8.9
        & 18.9\std7.9
        & 49.9\std8.6
        & 17.4\std4.9
        & 13.4\std3.0
        & 17.8\std3.5
        & 22.77
        \\
        \multirow{1}*{NNShot($\star$)}
        & 28.5\std9.2
        & 27.3\std12.3
        & 21.4\std9.7
        & 61.2\std10.4
        & 22.7\std7.4
        & 15.3\std1.6
        & 10.5\std2.9
        & 26.7
        \\
        \multirow{1}*{StructShot($\star$)}
        & 30.5\std12.3
        & 28.8\std11.2
        & 20.8\std9.9
        & 62.4\std10.5
        & 24.2\std8.0
        & 21.4\std3.8
        & 7.8\std2.1
        & 27.99
        \\
       \multirow{1}*{CONTaiNER($\star$)}
        & 32.2\std5.3
        & 30.9\std11.6
        & 32.9\std12.7
        & 57.8\std5.5
        & 24.2\std7.24
        & 16.4\std3.19
        & \textbf{17.9\std2.28}
        & 30.33
        \\
    \midrule
    \multirow{1}*{ProtoBERT($\dagger$)}
        & 8.39\std2.16	
        & 17.12\std4.04	
        & 8.4\std1.94	
        & 53.09\std9.89	
        & 21.17\std4.71
        & 15.85\std4.89
        & 11.91\std3.01
        & 19.42
        \\
    \multirow{1}*{NNShot($\dagger$)}
        & 21.97\std7.11
        & 33.89\std7.1
        & 21.73\std6.78
        & 59.76\std8.63
        & 26.53\std4.54
        & 15.0\std3.63	
        & 10.33\std3.08
        & 27.03
        \\
    \multirow{1}*{StructShot($\dagger$)}
        & 24.02\std6.24
        & 36.42\std8.22
        & 22.70\std6.65
        & 60.84\std7.62
        & 29.16\std4.88	
        & 18.34\std2.70
        & 11.17\std2.18
        & 28.95
        \\
    \multirow{1}*{CONTaiNER($\dagger$)}
        & 31.63\std11.74
        & 51.33\std8.97
        & 39.97\std3.81
        & 57.89\std16.79
        & 26.67\std8.65
        & 18.96\std3.97
        & 12.07\std1.53
        & 34.07
        \\
    \multirow{1}*{TransferBERT($\dagger$)}
        & 7.44\std5.97
        & 8.97\std4.94
        & 7.34\std3.42
        & 47.09\std11.02
        & 11.83\std5.07
        & \textbf{35.25\std4.21}
        & 8.97\std2.56
        & 18.13
        \\
    \multirow{1}*{DualEncoder($\dagger$)}
        & 0.83\std0.62
        & 2.86\std1.70
        & 2.55\std1.37
        & 54.63\std3.43
        & 36.03\std2.02
        & 14.63\std3.10
        & 11.87\std0.76
        & 17.63
        \\
    \multirow{1}*{EntLM($\dagger$)}
        & 5.79\std4.22
        & 10.11\std4.13
        & 8.49\std5.0
        & 50.47\std6.74
        & 27.7\std7.66
        & 7.85\std2.81
        & 8.85\std1.17
        & 17.04
        \\
    \multirow{1}*{DemonstrateNER($\dagger$)}
        & 0.98\std0.83
        & 2.02\std2.1
        & 4.02\std3.23
        & 16.12\std7.33
        & 20.38\std8.02
        & 13.29\std4.73
        & 3.24\std1.34
        & 8.58
        \\
    \multirow{1}*{\ours}
        & \textbf{37.94\std6.08}
        & \textbf{53.74\std3.6}
        & \textbf{46.27\std10.72}
        & \textbf{69.16\std4.47}
        & \textbf{43.89\std2.17}
        & 24.98\std3.44
        & 15.29\std1.89
        & \textbf{41.61}
        \\
    \midrule[1pt]
    \multicolumn{9}{c}{5-shot} \\
    \midrule[1pt]
    \multirow{1}*{ProtoBERT($\star$)}
            & 30.5\std3.5
            & 38.7\std5.6
            & 41.1\std3.3
            & 61.3\std9.1
            & 22.8\std4.5
            & 17.9\std1.8
            & 19.5\std3.4
            & 33.11
            \\
        \multirow{1}*{NNShot($\star$)}
            & 44.0\std2.1
            & 51.6\std5.9
            & 47.6\std2.8
            & 74.1\std2.3
            & 27.3\std5.4
            & 22.0\std1.5
            & 15.9\std1.8
            & 40.36
            \\
        \multirow{1}*{StructShot($\star$)}
            & 47.5\std4.0
            & 53.0\std7.9
            & 48.7\std2.7
            & 74.8\std2.4
            & 30.4\std6.5
            & 30.3\std2.1
            & 13.3\std1.3
            & 42.57
            \\
        \multirow{1}*{CONTaiNER($\star$)}
            & 51.2\std5.9
            & 55.9\std6.2
            & 61.5\std2.7
            & 72.8\std2.0
            & 27.7\std2.2
            & 24.1\std1.9
            & 24.4\std2.2
            & 45.37
            \\
    \midrule
    \multirow{1}*{ProtoBERT($\dagger$)}
        & 25.81\std3.0
        & 31.49\std4.6
        & 32.08\std2.12
        & 65.76\std5.34
        & 32.81\std8.78
        & 35.05\std12.25
        & 25.02\std2.66
        & 35.43
        \\
    \multirow{1}*{NNShot($\dagger$)}
        & 39.49\std5.96	
        & 50.18\std4.99	
        & 45.98\std4.61	
        & 70.79\std3.44	
        & 33.68\std5.21
        & 29.50\std2.89
        & 19.04\std2.38
        & 41.24
        \\
    \multirow{1}*{StructShot($\dagger$)}
        & 35.68\std6.17	
        & 51.30\std4.61
        & 47.85\std4.74	
        & 71.23\std3.62
        & 35.36\std2.99
        & 27.08\std3.17
        & 19.67\std2.45
        & 41.17
        \\
    \multirow{1}*{CONTaiNER($\dagger$)}
        & 45.62\std6.58
        & 67.70\std2.80
        & 59.84\std2.62
        & 75.48\std2.80
        & 35.83\std5.51
        & 30.14\std3.35
        & 16.19\std0.68
        & 47.26
        \\
    \multirow{1}*{TransferBERT($\dagger$)}
        & 21.48\std5.73
        & 41.97\std5.65
        & 45.24\std4.33
        & 69.93\std3.98
        & 35.64\std3.55
        & 47.89\std7.02
        & 27.50\std1.27
        & 41.38
        \\
    \multirow{1}*{DualEncoder($\dagger$)}
        & 7.61\std2.50
        & 16.41\std1.22
        & 26.37\std7.25
        & 67.05\std3.69
        & 36.82\std1.09
        & 23.27\std2.26
        & 24.55\std1.12
        & 28.87
        \\
    \multirow{1}*{EntLM($\dagger$)}
        & 21.29\std5.77
        & 35.7\std6.2
        & 28.8\std6.62
        & 60.58\std9.39
        & 30.26\std3.99
        & 13.51\std2.4
        & 13.35\std1.9
        & 29.07
        \\
    \multirow{1}*{DemonstrateNER($\dagger$)}
        & 49.25\std10.34
        & 63.02\std4.64
        & 61.07\std8.08
        & 73.13\std4.01
        & 43.85\std2.56
        & 36.36\std4.58
        & 18.01\std2.81
        & 49.24
        \\
    \multirow{1}*{\ours}
        & \textbf{52.46\std5.71}
        & \textbf{69.69\std2.19}
        & \textbf{67.58\std3.25}
        & \textbf{79.16\std4.49}
        & \textbf{53.41\std2.39}
        & \textbf{58.21\std3.58}
        & \textbf{36.99\std1.49}
        & \textbf{59.64}
        \\
    \bottomrule[1pt]
\end{tabular}
\end{threeparttable}
}
\label{tab:maintable1}
\end{table*}

\begin{table*}[t]
\setlength{\tabcolsep}{1.0mm}
\centering
\caption{\small Evaluation results of \ours and 7 baseline methods in episode evaluation protocol for FewNERD dataset. Results with $\star$ are reported by the original paper, and those with $\dagger$ are reproduced in our experiments. We report the averaged micro-F1 score together with standard deviation.}
\begin{threeparttable}
    \begin{tabular}{l|cc|cc|c}
    \toprule
    \multirow{2}{*}{Method}
    & \multicolumn{2}{c|}{1-shot} & \multicolumn{2}{c|}{5-shot} & \multirow{2}{*}{Avg.} \\
    & INTRA & INTER & INTRA & INTER  \\ 
    \midrule
    \multirow{1}*{ProtoBERT($\star$)}
            & 20.76
            & 38.83
            & 42.54
            & 58.79
            & 40.23
            \\
        \multirow{1}*{NNShot($\star$)}
            & 25.78
            & 47.24
            & 36.18
            & 55.64
            & 41.21
            \\
        \multirow{1}*{StructShot($\star$)}
            & 30.21
            & 51.88
            & 38.00
            & 57.32
            & 44.35
            \\
       \multirow{1}*{CONTaiNER($\star$)}
            & 40.43 
            & 53.70
            & 55.95
            & 61.83
            & 52.98
        \\
        \multirow{1}*{ESD($\star$)}
            & 36.08\std1.60 
            & 59.29\std1.25
            & 52.14\std1.50
            & 69.06\std0.80
            & 54.14
        \\
        \multirow{1}*{DecomposedMetaNER($\star$)}
            & 49.48\std0.85
            & 64.75\std0.35
            & 62.92\std0.57
            & 71.49\std0.47
            & 62.16
        \\
    \midrule
    \multirow{1}*{ProtoBERT($\dagger$)}
            & 25.8\std0.35	
            & 47.59\std0.84
            & 50.19\std0.65 
            & 65.05\std0.39
            & 47.16
            \\
    \multirow{1}*{NNShot($\dagger$)}
            & 33.32\std0.69	
            & 52.29\std0.88
            & 45.61\std0.52
            & 59.63\std0.48
            & 47.71
            \\
    \multirow{1}*{StructShot($\dagger$)}
        & 34.51\std0.68
        & 53.1\std0.92
        & 46.88\std0.48	
        & 60.45\std0.51
        & 48.74
        \\
    \multirow{1}*{CONTaiNER($\dagger$)}
        & 37.12\std1.01
        & 55.19\std0.43
        & 49.22\std0.34
        & 62.64\std0.33
        & 51.04
        \\
    \multirow{1}*{TransferBERT($\dagger$)}
        & 22.43\std1.49
        & 38.26\std2.36
        & 48.95\std1.23
        & 62.2\std1.36
        & 42.96
        \\
    \multirow{1}*{\ours}
        & \textbf{56.49\std0.56}
        & \textbf{68.23\std0.29}
        & \textbf{68.12\std0.35}
        & \textbf{75.3\std0.23}
        & \textbf{67.04}
        \\
    \bottomrule
\end{tabular}
\end{threeparttable}
\label{tab:maintable2}
\end{table*}

\section{Experiments}

\subsection{Setup}
\paragraph{Datasets}
We conduct experiments on multiple datasets across two few-shot NER formulations, tag-set extension and domain transfer. Following~\citet{das-etal-2022-container, yang-katiyar-2020-simple}, we split OntoNotes 5.0~\citep{weischedel2013ontonotes} into Onto-A, Onto-B, and Onto-C for the tag-set extension formulation.
For the domain transfer formulation, we use OntoNotes 5.0~\citep{weischedel2013ontonotes} as the source domain, CoNLL'03~\citep{DBLP:conf/conll/SangM03}, WNUT'17~\citep{DBLP:conf/aclnut/DerczynskiNEL17}, I2B2'14~\citep{DBLP:journals/jbi/StubbsU15}, and GUM~\citep{DBLP:journals/lre/Zeldes17} as target domains.
We also take Few-NERD~\citep{ding-etal-2021-nerd} as one of the tag-set extension tasks, which is a large-scale human-annotated dataset speciallly designed for few-shot NER.
We adopt the IO tagging scheme, where a label ``O'' is assigned to non-entity tokens and an entity type label is assigned to entity tokens.
We also transform the abbreviated label annotations into plain texts; e.g., [LOC] to [location].

\paragraph{Baselines}
Our baselines include metric learning based methods such as the prototypical networks ProtoBERT~\citep{protobert,DBLP:conf/sac/FritzlerLK19,hou-etal-2020-shot}, a nearest neighbor based network NNShot and its viterbi decoding variant StructShot ~\citep{yang-katiyar-2020-simple}, and a contrastive learning method CONTaiNER~\citep{das-etal-2022-container}. We also include a classification head based method TransferBERT~\citep{hou-etal-2020-shot} based on a pretrained BERT~\citep{devlin-etal-2019-bert}. Existing method that make use of label semantics, DualEncoder~\citep{ma-etal-2022-label} is also reproduced for comparison.
Recent prompt-based methods EntLM~\citep{ma2021template} and DemonstrateNER~\citep{lee-etal-2022-good} are also employed as the baselines as well. We also compare our model with the recently-introduced based methods DecomposeMetaNER~\citep{Ma2022DecomposedMF} and ESD~\citep{Wang2022AnES}. \footnote{The Few-NERD dataset has updated once. The dataset we used is Few-NERD Arxiv V6 Version, while ~\citet{Ma2022DecomposedMF, Wang2022AnES} reported their performances in the papers based on earlier version (i.e. Arxiv V5 Version). We find the performances on the latest Few-NERD dataset on their official github repo at \url{https://github.com/microsoft/vert-papers/tree/master/papers/DecomposedMetaNER}. }

\paragraph{Evaluation Protocols}
Following ~\citet{das-etal-2022-container, yang-katiyar-2020-simple}, we use the low-resource evaluation protocol for domain transfer tasks and for the tag-set extension tasks Onto-A, Onto-B, and Onto-C.
Since Few-NERD~\citep{ding-etal-2021-nerd} is specifically designed for episode evaluation, all of our experiments on Few-NERD dataset are evaluated under episode evaluation protocol. We follow the $N$-way $K$-shot downsampling setting proposed by ~\citet{ding-etal-2021-nerd}. It samples a set of sentences such that there are $N$ entity types in the sampled set with each entity type occurring $K\sim2K$ times in total, while removing any sentence from the set will make the previous conditions unsatisfied. For both evaluation protocols, we experiment with both $1$-shot and $5$-shot settings. 
For episode evaluation, we conduct $5$ different runs of experiments, each of them contains $5000$ test episodes. For low-resource evaluation, $10$ different runs of support set sampling is performed.

\begin{table*}[t]
\setlength{\tabcolsep}{3.5mm}
\centering
\small
\caption{\small Ablation Study for ProML. The variants \textbf{A} and \textbf{B} refer to using the option prefix prompt and label-aware prompt only, respectively. \textbf{Plain} refers to the original inputs and \textbf{plain+B} refers to replacing the option prefix prompt with the original inputs. \textbf{A+B} is our ProML method. 
}
\resizebox{\textwidth}{!}{%
\begin{threeparttable}
    \begin{tabular}{c|l|ccccc}
    \toprule[1pt]
    Setting & Model & Onto-A & Onto-B & Onto-C & INTRA & INTER \\ 
             

    \midrule[1pt]
    \multirow{9}*{\shortstack{5-shot}}
        & plain
            & 45.48\std1.03	
            & 64.21\std0.52
            & 51.26\std0.66
            & 49.22\std0.34
            & 62.64\std0.33
             
        \\
        & A (\mixrate$=1.0$)	
            & 45.91\std1.65
            & 60.06\std1.10
            & 51.02\std2.71
            & 65.64\std0.79
            & 74.22\std0.47 
        \\
        &  B (\mixrate$=0.0$)
            & 40.54\std2.04
            & 57.51\std0.70
            & 45.00\std3.34
            & 64.82\std0.55	
            & 72.74\std0.46
        
        \\
        & plain+B (\mixrate$=0.3$)	
            & 47.40\std0.54	
            & 59.08\std0.77
            & 53.33\std0.76
            & 64.21\std1.62
            & 73.15\std0.22
             
        \\
        & plain+B (\mixrate$=0.5$)	
            & 51.53\std0.47	
            & 63.19\std1.10
            & 57.03\std0.70
            & 64.54\std1.00	
            & 73.25\std0.44
            
        \\
        & plain+B (\mixrate$=0.7$)	
            & 51.12\std0.86
            & 65.47\std1.01
            & 58.38\std1.10
            & 63.18\std0.94	
            & 73.36\std0.67
            
        \\
        & A+B (\mixrate$=0.3$)	
            & 50.43\std0.82
            & 61.07\std1.50
            & 56.51\std1.06
            & 66.94\std0.56
            & 73.17\std0.34

        \\
        & A+B (\mixrate$=0.5$)	
            & 53.27\std0.62
            & 62.91\std0.91
            & 60.21\std1.25
            & 67.14\std0.71	
            & 74.26\std0.25
            
        \\
        & A+B (\mixrate$=0.7$)	
            & \textbf{54.56\std0.78}
            & \textbf{67.48\std0.64}	
            & \textbf{62.10\std1.01}
            & \textbf{68.12\std0.35}
            & \textbf{75.3\std0.23}
        \\
    \bottomrule[1pt]
\end{tabular}
\end{threeparttable}}
\label{tab:ablation}
\end{table*}

\begin{table*}[t]
\small
\caption{\small Case study: An illustration of some cases from the WNUT test set. There are $6$ entities: person (PER), location (LOC), product (PRO), creative work (CW), miscellaneous (MIS), group (GRO). Here {\color{blue}blue} color represents correct predictions, while {\color{red}red} color represents mistakes.}
\begin{tabularx}{\textwidth}{XXX}
\toprule
\textbf{GroundTruth} & \textbf{\ours} & \textbf{CONTaiNER} \\ \midrule
wow {\color{blue}emma$_{PER}$} and {\color{blue}kaite$_{PER}$} is so very cute and so funny i wish im {\color{blue}ryan$_{PER}$}&wow {\color{blue}emma$_{PER}$} and {\color{blue}kaite$_{PER}$} is so very cute and so funny i wish im {\color{blue}ryan$_{PER}$}               &wow {\color{blue}emma$_{PER}$} and {\color{blue}kaite$_{PER}$} is so very cute and so funny i wish {\color{red}im$_{PER}$} {\color{blue}ryan$_{PER}$}
                    \\ \midrule
great video ! good comparisons between the {\color{blue}ipad$_{PRO}$} and the {\color{blue}ipad$_{PRO}$} {\color{blue}pro$_{PRO}$} !&great video ! good comparisons between the {\color{blue}ipad$_{PRO}$} and {\color{red}the$_{PRO}$} {\color{blue}ipad$_{PRO}$} {\color{blue}pro$_{PRO}$} !&great video ! good comparisons between the {\color{red}ipad} and the {\color{red}ipad} {\color{blue}pro$_{PRO}$} !
\\
\midrule
i pronounce it nye-on cat&i pronounce it nye-on cat&i pronounce it {\color{red}nye-on$_{PRO}$} {\color{red}cat$_{PRO}$} \\
\bottomrule
\end{tabularx}
\label{tab:case_study}
\end{table*}

\paragraph{Training Details}
We use AdamW~\citep{DBLP:conf/iclr/LoshchilovH19} for optimization and the learning rate is set to $3\times 10^{-5}$, linearly warming up during first $10\%$ of all $10^4$ training iterations. The weight decay is set to $0.01$ for all parameters of the model except the biases and layer norm layers. The value of hyperparameter $\rho$ is chosen from $\{0.1, 0.3, 0.5, 0.7, 0.9\}$ and is set to $0.7$ by default (which is good enough for almost all cases). For fair comparison, we use the same Gaussian embedding dimension $d=128$ as CONTaiNER~\citep{das-etal-2022-container}. 


\subsection{Main Results}
The main results of low-resource evaluation and episode evaluation are shown in Tables \ref{tab:maintable1} and \ref{tab:maintable2} respectively.
We include both results reported by previous work and our reproduced results using the provided code.
Our method achieves new state-of-the-art (SOTA) results under 16 out of the 18 considered settings. To compare with previous SOTA across different settings, we collect the relative improvement fractions from all settings and then compute an average and a maximum over these fractions. The result shows that \ours substantially outperforming the previous SOTA by an average of 8.84\% and a maximum of 34.51\% (from 28\% to 37\% on GUM 5-shot) in relative gains of micro F1. These outstanding results show that our method is effective for few-shot NER tasks. EntLM~\citep{ma2021template} has poor performance, probably because it is sensitive to pivot word searching which highly depends on the dataset statistics. Compared with the other baselines, the performances of prompt-based baselines decrease by a larger margin in the 1-shot settings since they heavily rely on finetuning on support sets and get overfitting easily.

\paragraph{Training Curve Comparison}

Our architecture of using multiple prompts also mitigates overfitting. We conduct two experiments on Few-NERD to prove this empirically. Figure \ref{fig:visualizingoverfitting} demonstrates the training curves for CONTaiNER~\citep{das-etal-2022-container} and our model. From the curves we can see that the trends of performances over training set are similar while the performance of CONTaiNER on dev set stops increasing much earlier than \ours. Compared with CONTaiNER, our model gets much better in the later epochs. This shows that \ours suffers less from overfitting in the few-shot setting. 


\subsection{Ablation Study and Analysis}
The ablation study results on Onto-A, Onto-B, Onto-C, and Few-NERD are shown in Table \ref{tab:ablation}, \ref{tab:ablation_appendix}. 
We adopt the episode evaluation protocol due to its low variance.


\paragraph{Option Prefix Prompts \& Label-Aware Prompts}
According to Table \ref{tab:ablation}, overall, by comparing the best variant of prompting methods to ``plain'', using prompting consistently outperforms the methods without prompting. The improvements are consistent with our motivation in the earlier sections. With the help of label semantic annotations, the model is able to leverage this information to better learn the representation of each token. In addition, the model does not need to spend much capacity memorizing and inferring the underlying entity types for input tokens, which is crucial in the few-shot setting where labels are scarce.

On the Ontonotes datasets, the option prefix prompt achieves a better result than the label-aware prompt. On the Few-NERD dataset, it seems that the performances of these two prompts are close to each other. The difference might result from the way these datasets are constructed. Few-NERD is a large-scale and carefully annotated dataset which contains $66$ fine-grained entity types, while OntoNotes contains only $18$ coarse-grained entity types among which $7$ are numerical entity types. Considering that OntoNotes has less entity types with a large portion being numerical, directly adding label-aware prompts might not generalize well.
However, as we will show in the next section, combining the two prompts always leads to the best performance because the model is able to dynamically adapt to the two representations.


\paragraph{Effect of Masked Weighted Average}
According to Table \ref{tab:ablation}, 
with a properly selected averaging weight $\rho$, our \ours outperforms all baselines by a large margin among all tested datasets, which indicates that both prompts contribute to our final performance.
By adjusting the averaging weight $\rho$, we are able to balance the weights of the two representations for different data distributions. 
Importantly, $\rho = 0.7$ tends to work well in most of the settings, which can be used as the default hyperparameter in our framework without tuning. Moreover, even with other values of $\rho$, \ours still outperforms most of the ``plain'' baselines, demonstrating its robustness.

\begin{figure}[t]
    \centering
    \begin{minipage}[t]{0.48\textwidth}
    \centering
    \includegraphics[width=\textwidth]{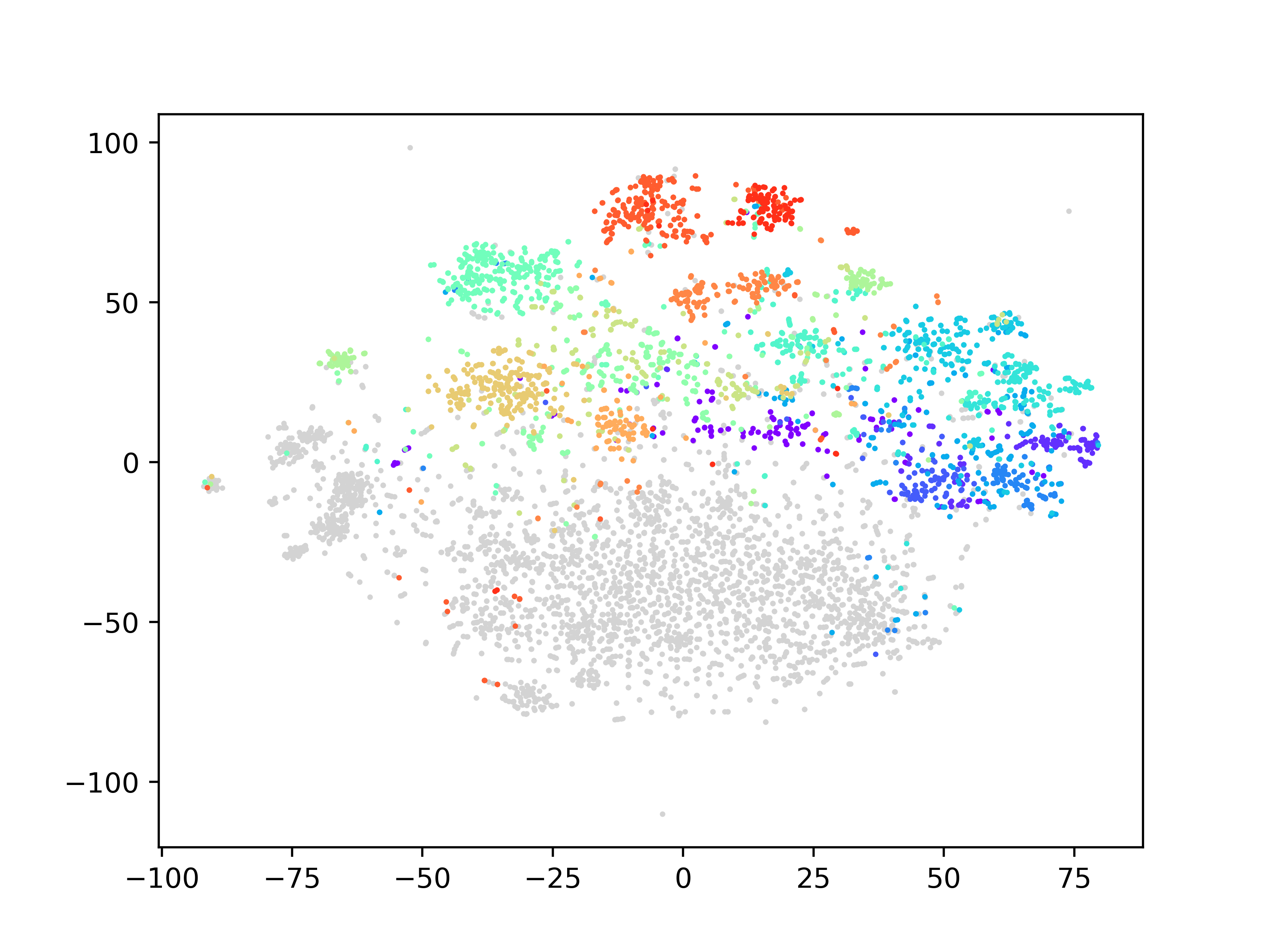}
    \end{minipage}
    \begin{minipage}[t]{0.48\textwidth}
    \centering
    \includegraphics[width=\textwidth]{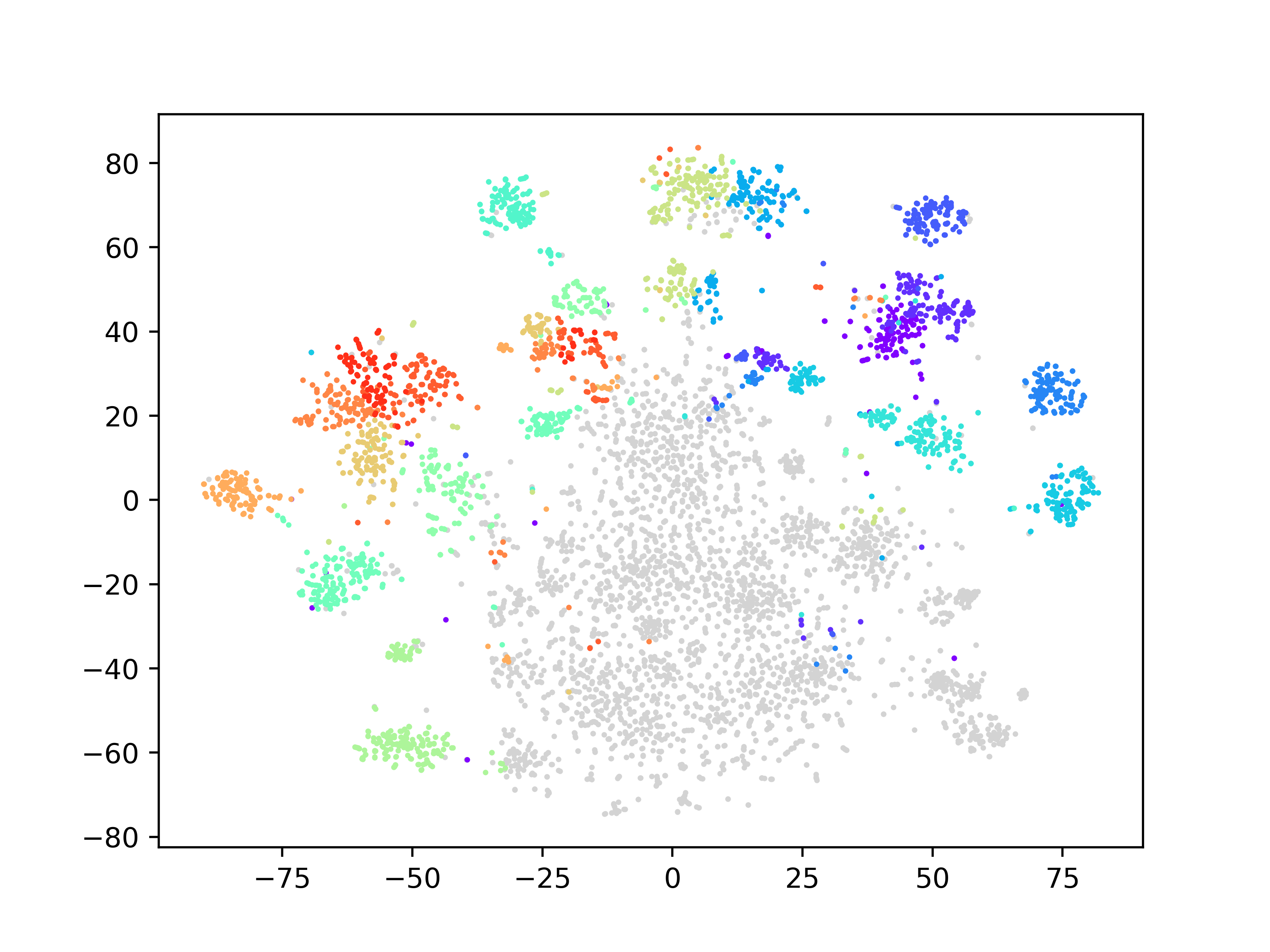}
    \end{minipage}
    \caption{\small TSNE visualization of token representations under the Few-NERD test set for CONTaiNER (on the left) and ProML (on the right), where each color represents an entity type (grey for non-entities). We only keep a fraction of $20\%$ among the non-entities to make the TSNE visualization clearer.}
    \label{fig:visualizingembeddings}
\end{figure}

\paragraph{Visualizing Embedding Space}
We visualize the token representations from support sets and query sets over several episodes from the test set of Few-NERD INTRA, as Figure~\ref{fig:visualizingembeddings} shows.
We observe that the token representations produced by \ours are concentrated in different clusters. In addition, we shall observe a clear decision boundary between different clusters. On the contrary, CONTaiNER seems to learn scattered, less separable features.

\paragraph{Case Study}

We present several randomly-selected cases from \ours and CONTaiNER using the test-set results of WNUT $1$-shot domain transfer task.
The results are in Table \ref{tab:case_study}.
We can see that \ours gives better predictions than CONTaiNER~\citep{das-etal-2022-container} for most cases. 
Specifically, CONTaiNER often misses entities or incorrectly classifies non-entities. 


\section{Conclusions}
We propose a novel prompt-based metric learning framework \ours for few-shot NER that leverages multiple prompts to guide the model with label semantics.
\ours is a general framework consistent with any token-level metric learning method and can be easily plugged into previous methods. We test \ours under 18 settings and find it substantially outperforms previous SOTA results by an average of 8.84\% and a maximum of 34.51\% in relative gains of micro F1. We perform ablation studies to show that multiple prompt schemas benefit the generalization ability for our model.

\bibliography{iclr2023_conference}
\bibliographystyle{iclr2023_conference}

\appendix
\section{Appendix}
\label{sec:appendix}
\subsection{Training Curve}
\begin{figure}[ht]
    \centering
    \begin{minipage}[t]{0.48\textwidth}
    \centering
    \includegraphics[width=\textwidth]{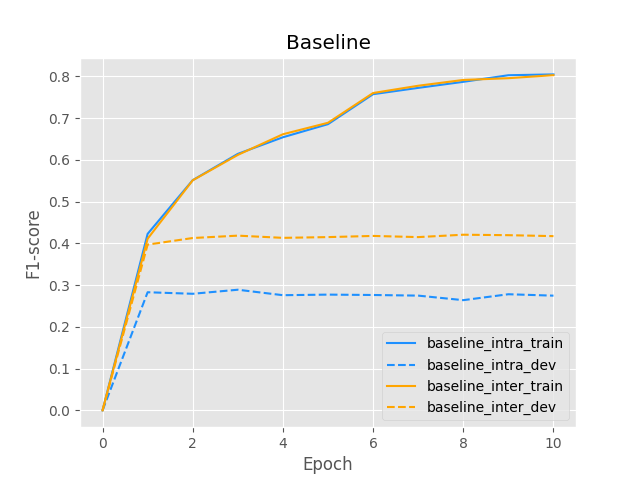}
    \end{minipage}
    \begin{minipage}[t]{0.48\textwidth}
    \centering
    \includegraphics[width=\textwidth]{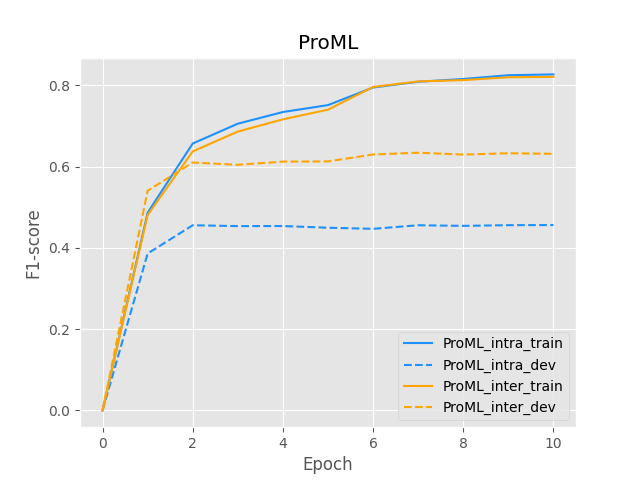}
    \end{minipage}
    \caption{\small Training curves for CONTaiNER~\cite{das-etal-2022-container} baseline (on the left) and our model (on the right). The experiments are conducted under Few-NERD INTRA 1-shot and INTER 1-shot setting.}
    \label{fig:visualizingoverfitting}
\end{figure}
\subsection{Ablation for 1-shot}
\begin{table*}[ht]
\setlength{\tabcolsep}{3.5mm}
\centering
\small
\caption{\small Ablation Study for ProML (1-shot and 5-shot). \textbf{A, B} refer to the option prefix prompt and label-aware prompt, respectively. \textbf{Plain} refers to the original inputs and \textbf{plain+B} refers to replacing the option prefix prompt with the original inputs. All results in this table are produced by the episode evaluation protocol.
}
\resizebox{\textwidth}{!}{%
\begin{threeparttable}
    \begin{tabular}{c|l|ccccc}
    \toprule[1pt]
    Setting & Model & Onto-A & Onto-B & Onto-C & INTRA & INTER \\ 
    \midrule[1pt]
    \multirow{8}*{\shortstack{1-shot}}
    & plain
            & 27.43\std0.54	
            & 48.89\std1.1
            & 33.54\std1.3
            & 37.12\std1.01
            & 55.19\std0.43
             
        \\
    & A (\mixrate$=1.0$)	
        & 32.47\std1.21	
        & 46.18\std3.27	
        & 34.21\std1.11	
        & 50.01\std1.36	
        & 65.07\std0.05
        \\
    & B (\mixrate$=0.0$)	
        & 28.56\std1.37
        & 40.08\std2.18
        & 28.00\std0.91
        & 51.8\std0.8
        & 66.44\std1.27
        \\
    & plain+B (\mixrate$=0.3$) 
        & 31.45\std1.88
        & 41.42\std0.96
        & 35.40\std0.73
        & 53.31\std0.79
        & 67.02\std1.2
        \\
    & plain+B (\mixrate$=0.5$)
        & 34.19\std1.26
        & 46.42\std1.92
        & 37.31\std1.15
        & 52.69\std1.19
        & 67.03\std1.07
        \\
    & plain+B (\mixrate$=0.7$)
        & 33.69\std0.83
        & 49.75\std0.50
        & 39.15\std0.37
        & 50.11\std0.57
        & 66.8\std0.62
        \\
    & ProML (\mixrate$=0.3$)	
        & 33.31\std0.51	
        & 43.13\std1.61	
        & 36.91\std2.15	
        & 55.01\std0.52	
        & 67.26\std0.42
        \\
    & ProML (\mixrate$=0.5$)	
        & 34.06\std0.74
        & 45.86\std1.27
        & 40.75\std1.14
        & \textbf{56.49\std0.56}
        & 67.97\std0.40
        \\
    & ProML (\mixrate$=0.7$)	
        & \textbf{34.82\std0.98}
        & \textbf{50.41\std0.87}
        & \textbf{42.85\std0.91}
        & 55.74\std0.61
        & \textbf{68.23\std0.29}
        \\

    \midrule[1pt]
    \multirow{9}*{\shortstack{5-shot}}
        & plain
            & 45.48\std1.03	
            & 64.21\std0.52
            & 51.26\std0.66
            & 49.22\std0.34
            & 62.64\std0.33
             
        \\
        & A (\mixrate$=1.0$)	
            & 45.91\std1.65
            & 60.06\std1.10
            & 51.02\std2.71
            & 65.64\std0.79
            & 74.22\std0.47 
        \\
        &  B (\mixrate$=0.0$)
            & 40.54\std2.04
            & 57.51\std0.70
            & 45.00\std3.34
            & 64.82\std0.55	
            & 72.74\std0.46
        
        \\
        & plain+B (\mixrate$=0.3$)	
            & 47.40\std0.54	
            & 59.08\std0.77
            & 53.33\std0.76
            & 64.21\std1.62
            & 73.15\std0.22
             
        \\
        & plain+B (\mixrate$=0.5$)	
            & 51.53\std0.47	
            & 63.19\std1.10
            & 57.03\std0.70
            & 64.54\std1.00	
            & 73.25\std0.44
            
        \\
        & plain+B (\mixrate$=0.7$)	
            & 51.12\std0.86
            & 65.47\std1.01
            & 58.38\std1.10
            & 63.18\std0.94	
            & 73.36\std0.67
            
        \\
        & ProML (\mixrate$=0.3$)	
            & 50.43\std0.82
            & 61.07\std1.50
            & 56.51\std1.06
            & 66.94\std0.56
            & 73.17\std0.34

        \\
        & ProML (\mixrate$=0.5$)	
            & 53.27\std0.62
            & 62.91\std0.91
            & 60.21\std1.25
            & 67.14\std0.71	
            & 74.26\std0.25
            
        \\
        & ProML (\mixrate$=0.7$)	
            & \textbf{54.56\std0.78}
            & \textbf{67.48\std0.64}	
            & \textbf{62.10\std1.01}
            & \textbf{68.12\std0.35}
            & \textbf{75.3\std0.23}
        \\
    \bottomrule[1pt]
\end{tabular}
\end{threeparttable}}
\label{tab:ablation_appendix}
\end{table*}

\end{document}